# Bayesian Hyperparameter Optimization with BoTorch, GPyTorch and Ax

Daniel T. Chang (张遵)

*IBM (Retired)* dtchang43@gmail.com

## Abstract:

Deep learning models are full of hyperparameters, which are set manually before the learning process can start. To find the best configuration for these hyperparameters in such a high dimensional space, with time-consuming and expensive model training / validation, is not a trivial challenge. Bayesian optimization is a powerful tool for the joint optimization of hyperparameters, efficiently trading off exploration and exploitation of the hyperparameter space. In this paper, we discuss Bayesian hyperparameter optimization, including hyperparameter optimization, Bayesian optimization, and Gaussian processes. We also review BoTorch, GPyTorch and Ax, the new open-source frameworks that we use for Bayesian optimization, Gaussian process inference and adaptive experimentation, respectively. For experimentation, we apply Bayesian hyperparameter optimization, for optimizing group weights, to weighted group pooling, which couples unsupervised tiered graph autoencoders learning and supervised graph prediction learning for molecular graphs. We find that Ax, BoTorch and GPyTorch together provide a simple-to-use but powerful framework for Bayesian hyperparameter optimization, using Ax's high-level API that constructs and runs a full optimization loop and returns the best hyperparameter configuration.

## 1 Introduction

Deep learning models are full of *hyperparameters* (e.g., learning rate, number of layers, number of units per layer), which are set manually before the learning process can start. To find the best configuration for these hyperparameters in such a high dimensional space, with time-consuming and expensive model training / validation, is not a trivial challenge [4-6]. There are four common methods for *hyperparameter optimization*, in order of increasing efficiency: manual, grid search, random search, and Bayesian optimization.

*Bayesian optimization* [7-11], a sequential model-based optimization, is a powerful tool for the joint optimization of hyperparameters, efficiently trading off exploration and exploitation of the hyperparameter space. It is best-suited for optimization over continuous domains of less than 20 hyperparameters, and tolerates stochastic noise in function evaluations.

In this paper, we discuss *Bayesian hyperparameter optimization*, including *hyperparameter optimization*, *Bayesian optimization*, and *Gaussian processes* [12-13]. Our focus is on providing a concise and consistent description, highlighting essential aspects and mathematical representations.

We also review *BoTorch* [14-15], *GPyTorch* [16-17] and *Ax* [18-20], the new open-source frameworks, built on top of PyTorch, that we use for Bayesian optimization, Gaussian process inference and adaptive experimentation (e.g., Bayesian hyperparameter optimization), respectively. BoTorch supports seamless integration with GPyTorch, and is best used in tandem with Ax.

For experimentation, we apply Bayesian hyperparameter optimization to *weighted group pooling* for *optimizing group weights*. *Tiered graph autoencoders* [1-2] and *graph prediction* together provide effective, efficient and interpretable *deep learning for molecular graphs* [3], with the former providing unsupervised, transferable learning and the latter providing supervised, task-optimized learning. Tiered graph autoencoders learning and graph prediction learning are essentially separated, coupled only through weighted group pooling [3], which are parameterized by group weights.

## 1.1 Notations and Symbols

*Vectors* are in bold type. *Matrices* are capitalized and in bold type.

| ***Symbol*** | ***Meaning*** |
|---|---|
| $\sim$ | distributed according to |
| $\alpha(\mathbf{x})$ | *acquisition function* |
| d | dimension of *hyperparameter space* |
| $\mathcal{D}$ | data set: $D = \{(\mathbf{x}_i, y_i) \mid i = 1, \ldots, N\}$ |
| $\mathbb{E}[]$ | expectation |
| $f(\mathbf{x})$ or $\mathbf{f}$ | *objective function* or objective function values, $\mathbf{f} = [f(\mathbf{x}_1), \ldots, f(\mathbf{x}_N)]$ |
| $\mathcal{GP}$ | *Gaussian process*: $f(\mathbf{x}) \sim \mathcal{GP}(m(\mathbf{x}), k(\mathbf{x}, \mathbf{x}'))$ |
| $k(\mathbf{x}, \mathbf{x}')$ | *kernel (or covariance) function* evaluated at $\mathbf{x}$ and $\mathbf{x}'$ |
| $\mathbf{K}$ or $\mathbf{K}(\mathbf{X}, \mathbf{X})$ | N x N covariance matrix |
| $m(\mathbf{x})$ or $\mu(\mathbf{x})$ | the *mean function* of a Gaussian process |
| $\boldsymbol{\mu}$ or $\boldsymbol{\mu}(\mathbf{X})$ | N mean vector |
| $\mathcal{N}(\boldsymbol{\mu}, \mathbf{K})$ | *Gaussian (Normal) distribution* with mean vector $\boldsymbol{\mu}$ and covariance matrix $\mathbf{K}$ |
| N | number of training (and test) cases |
| $y \mid x$ and $p(y \mid x)$ | conditional random variable y given x and its *probability distribution* |
| R | the real numbers |
| $\sigma_\epsilon^2$ | noise variance |
| $\boldsymbol{\theta}$ | vector of *Gaussian process parameters* (parameters of the kernel function) |
| $\mathcal{X}$ | *feasible set* (input space) and also the index set for the stochastic process |
| $\mathbf{X}$ | N × d matrix of the training inputs $\{\mathbf{x}_i\}$, i =1 to N |
| $\mathbf{x}_i$ | the *i*th input (vector of *hyparameters*) |
| $x_j$ | the *j*th *hyperparameter*, j = 1 to d |
| $y_i$ | the *i*th observation: $y_i \sim \mathcal{N}(f(\mathbf{x}_i), \sigma_\epsilon^2)$ |

## 2 Hyperparameter Optimization

Deep learning models are full of hyperparameters (e.g., learning rate, number of layers, number of units per layer), which are set manually before the learning process can start. To find the best configuration for these hyperparameters in such a high dimensional space, with time-consuming and expensive model training / validation, is not a trivial challenge [4-6].

Hyperparameter optimization is represented in mathematical form as [5]:

$$\mathbf{x}_{opt} = \text{argmin } f(\mathbf{x}), \ \mathbf{x} \in \mathcal{X}$$

where $\mathbf{x}$ is the vector of hyperparameters, $\mathcal{X}$ is the *feasible set (domain)*, and $f(\mathbf{x})$ is the *objective score* to minimize, evaluated on the *validation set*. In simple terms, the goal is to find the hyperparameters that yield the best score on the *validation set metric*. The problem with hyperparameter optimization is that evaluating the *objective function f* to find the score is extremely time-consuming and costly with a large number of hyperparameters and complex models, such as deep learning models, that involve time-consuming and expensive *train-predict-evaluate cycles*.

There are four common methods [5-6] for hyperparameter optimization, in order of increasing efficiency:

- Manual
- Grid search
- Random search
- Bayesian optimization

The manual method does not scale. Grid search is infeasible with more than 4 hyperparameters. Random search (and grid search) has the downside that each new guess is independent of the previous iteration and, therefore, the search is incapable of leveraging learning for improvement.

*Bayesian optimization*, on the other hand, efficiently trades off exploration and exploitation of the hyperparameter space to quickly guide the search into the configuration that best optimizes some overall evaluation criterion [10-11]. Bayesian optimization is a *sequential model-based optimization (SMBO)* [6, 8, 10], which has five key aspects:

1. A *domain (input space)* of hyperparameters.
2. An *objective function* which takes in hyperparameters and outputs a *score* that we want to minimize.
3. A *surrogate model* of the objective function.
4. A *select function* for evaluating which hyperparameters to select next from the surrogate model.
5. A history consisting of *(hyperparameters, score) pairs* used to update the surrogate model.

## 3 Bayesian Optimization

*Bayesian optimization (BO)* [7-11] is a powerful tool for the joint optimization of hyperparameters, efficiently trading off exploration and exploitation of the hyperparameter space. It is best-suited for optimization over continuous domains of less than 20 hyperparameters, and tolerates stochastic noise in function evaluations.

BO focuses on solving the problem:

$$\text{(max or min) } f(\mathbf{x}), \quad \mathbf{x} \in \mathcal{X}$$

where $\mathbf{x}$ is the input (e.g., hyperparameters), $\mathcal{X}$ is the *feasible set (domain)*, and f is the *objective function*. It is designed for *data-efficient global optimization* with the following properties [9]:

- $\mathbf{x} \in \mathrm{R}^d$ for a value of d that is not too large. Typically $d < 20$.
- $\mathcal{X}$ is a simple set. Typically, $\mathcal{X}$ is a hyper-rectangle: $a_j <= x_j <= b_j$.
- f is *continuous*.
- f is *time-consuming and expensive* to evaluate, a *black-box*, and *derivative-free*.
- f may be obscured by *stochastic noise*.

BO consists of two main components: a *probabilistic (surrogate) model* for modeling the objective function, and an *acquisition function* that encodes a strategy for navigating the exploration vs. exploitation trade-off of the input space. We prescribe the *Bayesian prior*, $p(f(\mathbf{x}))$, a *prior belief (model)* over the possible objective functions and then sequentially refine this model as data are observed via Bayesian posterior updating. The *Bayesian posterior*, $p(f(\mathbf{x})|\mathcal{D})$, represents our *updated belief (model) – given data –* on the likely objective functions we are optimizing. Given this probabilistic model, we can *sequentially induce acquisition functions* that leverage the uncertainty in the posterior to guide exploration. In short, BO builds a probabilistic model of the objective function and uses it to acquire the most promising input data to evaluate the true objective function. Our only recourse is to *evaluate f at a sequence of inputs*, with the hope of determining a *near-optimal value* after a small number of evaluations.

## 3.1 Probabilistic (Surrogate) Models

For *continuous objective functions*, *Bayesian optimization* typically works by assuming the probabilistic model is a *Gaussian process (prior)* and maintains a *posterior* distribution as the results of evaluating objective functions are observed. Other choices [10] for the probabilistic model include random forests and tree Parzen estimators (TPEs). Gaussian processes are flexible and powerful.

The *Gaussian process (GP)* [7] is a convenient and powerful prior distribution on *unknown functions* of the form

$f : \mathcal{X} \to \mathrm{R}$

It is defined by the property that any finite set of N points

$\{\mathbf{x}_i \in \mathcal{X}\}$, i = 1 to N

induces a *multivariate Gaussian distribution* on $\mathrm{R}^N$. The *i*th of these points is taken to be the function value $f(\mathbf{x}_i)$. We discuss Gaussian processes in Section 4 Gaussian Processes.

## 3.2 Acquisition Functions

We assume [7] that the objective function $f(\mathbf{x})$ is drawn from a surrogate *Gaussian process prior*, as discussed above, and that our *observations (history)* are of the form

$\{\mathbf{x}_i, y_i\}$, i = 1 to N

where

$y_i \sim \mathcal{N}(f(\mathbf{x}_i), \sigma_\epsilon^2)$

and $\sigma_\epsilon^2$ is the *variance of noise* introduced into the observations. This prior and these observations induce a *posterior* over functions.

We denote the *acquisition function* [7] by

$\alpha : \mathcal{X} \rightarrow R^{+}$

It determines what point in $\mathcal{X}$ should be selected next via a proxy optimization:

$\mathbf{x}_{next} = \text{argmax } \alpha(\mathbf{x}), \mathbf{x} \in \mathcal{X}$

In general, acquisition functions depend on the *previous observations*, as well as the *GP parameters* $\boldsymbol{\theta}$ (parameters of the GP kernel function, see 4 Gaussian Processes). We denote this dependence as

$\alpha(\mathbf{x} \mid \{\mathbf{x}_i, y_i\}, \boldsymbol{\theta})$.

Under the Gaussian process prior, acquisition functions depend on the model solely through its *mean function*

$\mu(\mathbf{x} \mid \{\mathbf{x}_i, y_i\}, \boldsymbol{\theta})$

and *variance function*

$\sigma^2(\mathbf{x} \mid \{\mathbf{x}_i, y_i\}, \boldsymbol{\theta})$.

Acquisition functions define a balance [10] between exploring new areas in the objective function space and exploiting areas that are already known to have favorable values. A common strategy is to maximize the *expected improvement (EI)* over the current best result. This has closed form under the Gaussian process [7]:

$\alpha_{EI}(\mathbf{x} \mid \{\mathbf{x}_i, y_i\}, \boldsymbol{\theta}) = \sigma(\mathbf{x} \mid \{\mathbf{x}_i, y_i\}, \boldsymbol{\theta})(\gamma(\mathbf{x})\Phi(\gamma(\mathbf{x})) + \mathcal{N}(\gamma(\mathbf{x}) \mid 0, 1)$.

Other strategies include probability improvement and GP upper confidence bound. EI is better-behaved than probability of improvement and, unlike GP upper confidence bound, it does not require its own tuning parameters.

# 4 Gaussian Processes

The *Gaussian process* [12-13] is a well-known *non-parametric* and interpretable Bayesian probabilistic model. A Gaussian process is a generalization of the *Gaussian distribution*. Whereas a probability distribution describes random variables which are scalars or vectors, a *stochastic process* governs the properties of *functions*. We can loosely think of a function as *an infinite vector*, each entry in the vector specifying the function value $f(\mathbf{x}_i)$ at a particular input $\mathbf{x}_i$. A key aspect of this is that if we ask only for the properties of the function at *a finite number of points*, then *GP inference* will give us the same answer if we ignore the infinitely many other points.

In the *function-space view* [12] a Gaussian process defines a *distribution over functions*, and inference takes place directly in the *space of functions*. A Gaussian process is completely specified by its *mean function* (average of all functions) and *kernel (covariance) function* (how much individual functions can vary around the mean function):

$$m(\mathbf{x}) = \mathbb{E}[f(\mathbf{x})],$$

$$k(\mathbf{x}, \mathbf{x}') = \mathbb{E}[(f(\mathbf{x}) - m(\mathbf{x}))(f(\mathbf{x}') - m(\mathbf{x}'))]$$

As such, we write the Gaussian process as

$$f(\mathbf{x}) \sim \mathcal{GP}(m(\mathbf{x}), k(\mathbf{x}, \mathbf{x}')).$$

Under the Gaussian process, the true objective is modeled by a *GP prior* with a mean and a kernel function. Given a set of (noisy) *observations* from initial evaluations, a Bayesian posterior update gives the *GP posterior* with an *updated mean and kernel function*. The mean function of the GP posterior gives the best *predictions* at any point conditional on the available observations, and the kernel function quantifies the *uncertainty* in the predictions. The *GP prior* is a multivariate Gaussian distribution:

$$p(\mathbf{f} \mid \boldsymbol{\theta}) \sim \mathcal{N}(\boldsymbol{\mu}, \mathbf{K}(\boldsymbol{\theta})).$$

So is the GP posterior:

$$p(\mathbf{f} \mid \mathcal{D}, \boldsymbol{\theta}) \sim \mathcal{N}(\boldsymbol{\mu}', \mathbf{K}'(\boldsymbol{\theta})).$$

In the above, $\boldsymbol{\theta}$ is *GP parameters* (parameters of the GP kernel function).

# 5 BoTorch, GPyTorch and Ax

*BoTorch*, *GPyTorch*, and *Ax* are new open-source frameworks, built on top of PyTorch, for Bayesian optimization, Gaussian process inference, and adaptive experimentation (e.g., Bayesian hyperparameter optimization), respectively. BoTorch supports seamless integration with GPyTorch, and is best used in tandem with Ax.

## 5.1 BoTorch

*BoTorch* [14-15] is a scalable framework for *Bayesian optimization*, enabled by *analytic* and *Monte-Carlo (MC) acquisition functions* and *auto-differentiation*. Its modular design facilitates flexible specification and optimization of probabilistic models, and simplifies implementation of novel acquisition functions. BoTorch provides seamless integration with PyTorch modules, enabling *joint training of GP and neural network modules* and allowing end-to-end *gradient-based optimization* of acquisition functions operating on differentiable models.

BoTorch provides abstractions for (combining) BO primitives, enabling BO with auto-differentiation, automatic parallelization, device-agnostic hardware acceleration, and generic neural network operators and modules. BoTorch consists of the following *abstractions for BO primitives*: Model, AcquisitionFunction, Objective, and Optimizer, as shown [14] and discussed below:

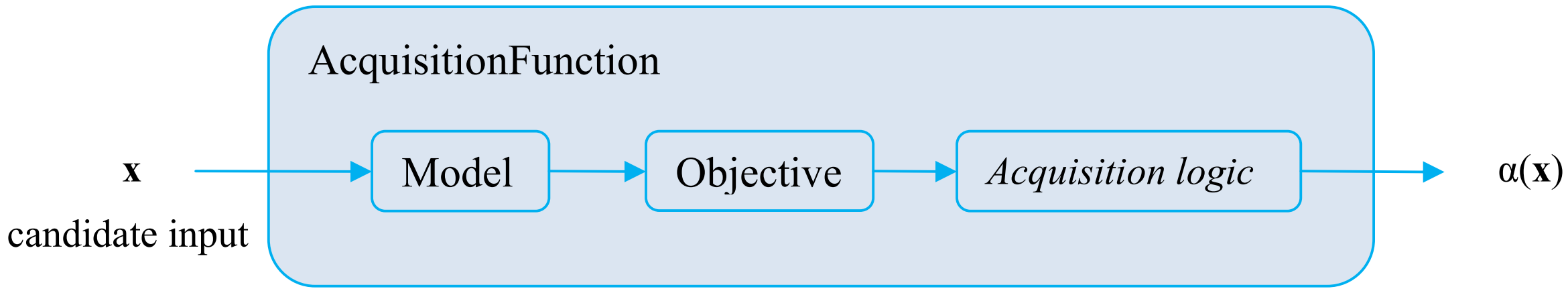

### *Model*

The *Model* is an abstraction for a probabilistic (surrogate) model (see 3.1 Probabilistic (Surrogate) Models). A Model maps a set of inputs to a *posterior distribution* of its outputs. It requires only a single *posterior()* method that returns a *Posterior* object describing the posterior distribution.

BoTorch is model-agnostic — the only requirement for a model is that, given a set of inputs, it can produce *posterior draws* of one or more outputs. *Explicit posteriors*, such as those provided by a GP, can be used directly. BoTorch provides a simple *Posterior* API that only requires implementing an *rsample()* method for sampling from the posterior. A Posterior can be *any distribution (even an implicit one)*, so long as one can sample from that distribution.

BoTorch provides first-class support for state-of-the-art probabilistic models in *GPyTorch* (see 5.2 GPyTorch). This includes support for multi-task GPs, deep kernel learning, deep GPs, and approximate inference. BoTorch provides several GPyTorch models to cover most standard BO use cases. *GPyTorchModel* provides a base class for conveniently wrapping GPyTorch models. For these models, *GPyTorchPosterior* should be used.

### *AcquisitionFunction*

The *AcquisitionFunction* is an abstraction for the acquisition function (see 3.2 Acquisition Functions). It implements a forward pass that takes a candidate input $\mathbf{x}$ and computes their acquisition function $\alpha(\mathbf{x})$. *Analytic AcquisitionFunctions* operate on the explicit posterior (e.g., GP), whereas *MC-based AcquisitionFunctions* operate on samples from the posterior evaluated under the Objective.

The idea behind MC-based AcquisitionFunctions is simple: instead of computing an (intractable) expectation over the posterior, we sample from the posterior and use the *sample average* as an approximation. All MC-based AcquisitionFunctions are derived from *MCAcquisitionFunction*. Any *Posterior* object can be used with an MCAcquisitionFunction.

### *Objective*

The *Objective* allows for convenient *transformation* of Model outputs into a scalar function to be optimized. All Objectives are derived from *MCAcquisitionObjective*. BoTorch implements several MC-based Objectives, including LinearMCObjective for linear combinations of model outputs, and ConstrainedMCObjective for constrained objectives.

### *Optimizer*

The *optimizer* finds

$\mathbf{x}_{\text{next}} = \text{argmax}\ \alpha(\mathbf{x}),\ \mathbf{x} \in \mathcal{X}$

in order to proceed with the next iteration of BO. *Auto-differentiation* makes it straightforward to use *gradient-based optimization*, which typically performs better than derivative-free approaches.

Optimization over N inputs in a d-dimensional hyperparameter space results in Nd scalar hyperparameters. Both N and d are often small compared to deep learning inputs and model parameters, respectively. As such, BoTorch provides a custom interface that wraps the *optimizers from the scipy.optimize module*, since these can be used but are not in the torch.optim module.

## 5.2 GPyTorch

*GPyTorch* [16-17] is a framework for *scalable GP inference* and *Bayesian deep learning*. GPyTorch uses *Blackbox Matrix-Matrix multiplication (BBMM)* for GP inference. BBMM reduces the asymptotic complexity of GP inference from $\mathcal{O}(n^3)$ to $\mathcal{O}(n^2)$. In addition, BBMM effectively uses GPU hardware to accelerate both GP inference and scalable approximations.

GPytorch supports both exact GP inference and variational GP inference. Exact GP inference is supported via the *ExactGP* model; variational GP inference is supported via the *ApproximateGP* model. The easiest way to use the *BoTorch GPyTorchModel*, discussed earlier, is to subclass a model from it and a GPyTorch model (e.g. an ExactGP). Such is the case for the *BoTorch SingleTaskGP* model, which works with independent output(s) and all outputs using the same training data, as well as the *BoTorch Fixed NoiseGP* model, a single-task exact GP that uses fixed observation noise levels.

## 5.3 Ax

*Ax* [18-20] is a framework for *adaptive experimentation* that automates the process of *sequential optimization* (e.g., *Bayesian optimization*). It provides an easy-to-use interface for defining, managing and running sequential experiments, while handling metadata management, transformations, and systems integration.

### *Components*

In Ax, an *Experiment* keeps track of the whole optimization process. It contains a search space, optimization configuration, metadata, information on what metrics to track, and how to run iterations, etc.

A *SearchSpace* is composed of a set of *Parameters* to be optimized in the Experiment, and optionally a set of parameter constraints. Each Parameter has a name, a type (int, float, bool, or string), and a domain. There are three kinds of Parameter: *RangeParameter*, ChoiceParameter and FixedParameter.

An *Arm* is a set of Parameters and their values with a name attached to it. In the case of hyperparameter optimization, an Arm corresponds to a *hyperparameter configuration* explored in the course of a given optimization.

An experiment consists of a sequence of *Trials*, each of which *evaluates* one or more Arms. Based on the evaluation results, the optimization algorithm *suggests* one or more Arms to evaluate. A Trial is added to the Experiment when a new Arm (or Arms) is proposed by the optimization algorithm. A Trial goes through multiple phases during the experimentation cycle, tracked by its *TrialStatus* field.

An *OptimizationConfig* is composed of an *Objective* with *Metric* to be minimized or maximized, and optionally a set of outcome constraints. The *Metric* provides an interface for *fetching data* for a Trial. All Metric classes must implement the method *fetch_trial_data()*, which accepts a Trial and returns an instance of *Data*. Each row of the final Data object represents the evaluation of an Arm on a Metric.

### Models

The *Model* represents a *probabilistic (surrogate) model* and predicts the outcomes of Metrics evaluated at an Arm. All Models share a common API with *predict()* to make predictions at new Arms and *gen()* to generate candidate Arms to be evaluated. Models are created using factory functions from the *ax.modelbridge.factory*. In particular, the *get_botorch()* function instantiates a *BotorchModel*, to be discussed later.

All Models can be used with the *built-in plotting utilities*, which can produce plots of model predictions on 1-d or 2-d slices of the parameter space. Ax also includes *utilities for cross validation* to assess model predictive performance.

Ax uses a bridge design to provide a unified interface for models. The modeling stack consists of two layers: the ModelBridge and the Model. The *ModelBridge* is the object that is directly used in Ax: model factories return ModelBridge objects, and plotting and cross validation tools operate on a ModelBridge. Model objects are only used via a ModelBridge. The primary role of the ModelBridge is to act as a *transformation layer*. This includes transformations to the data, search space, and optimization configuration, as well as the final transform from Ax objects into the objects consumed by the Model.

### *Ax with BoTorch*

Ax relies on *BoTorch* for implementing *Bayesian optimization* algorithms. It provides a *BotorchModel* that is a default for modeling and optimization, which can be customized by specifying and passing in *model constructors, acquisition functions, and optimization strategies*. This *TorchModelBridge*, for BotorchModel, utilizes a number of built-in *transformations*, such as normalizing inputs and outputs to ensure reasonable fitting of GPs. *GPs* are used for Bayesian optimization in Ax. The *get_GPEI()* function constructs a model that fits a GP to the data, and uses the BoTorch *ExpectedImprovement* acquisition function to generate new points.

### *The optimize() Function*

Ax provides a simple-to-use but powerful API for Bayesian hyperparameter optimization, *optimize*(), which constructs and runs a full *OptimizationLoop*, given *hyperparameters* and an *evaluation function*, and returns the *best hyperparameter configuration*. The OptimizationLoop creates a *SimpleExperiment* with the evaluation function and an associated *SearchSpace* with the hyperparameters. By default, the OptimizationLoop uses *20 Trials* per SimpleExperiment and *1 Arm* per Trial.

When a SimpleExperiment is constructed, it creates an *OptimizationConfig* with an *Objective* and a *Metric*. To run a trial, a SimpleExperiment first fetches the trial data (an instance of *Data*) for the previous existing Trial, and creates a *new Trial* with the trial data and pending observations. The generation strategy uses *Sobol* for the first 5 Arms and *GPEI* for subsequent Arms. Iterations after 5 will take longer to generate due to model-fitting.

In addition to the best hyperparameter configuration, the optimize() function also returns the *TorchModelBridge* used, with its associated *BotorchModel*. By default, this model uses a noisy *ExpectedImprovement* acquisition function on top of a model made up of separate GPs, one for each outcome.

# 6 Bayesian Hyperparameter Optimization Applied to Weighted Group Pooling

## 6.1 Tiered Graph Autoencoders with Graph Prediction

*Tiered graph autoencoders* [1-2] provide the most direct and effective architecture and mechanisms for generating *hierarchical representations (embeddings) of molecular graphs* since they are based on the direct representation and utilization of *groups*. With tiered graph autoencoders, we use *tiered graph embeddings* for molecular *graph prediction*, as shown below:

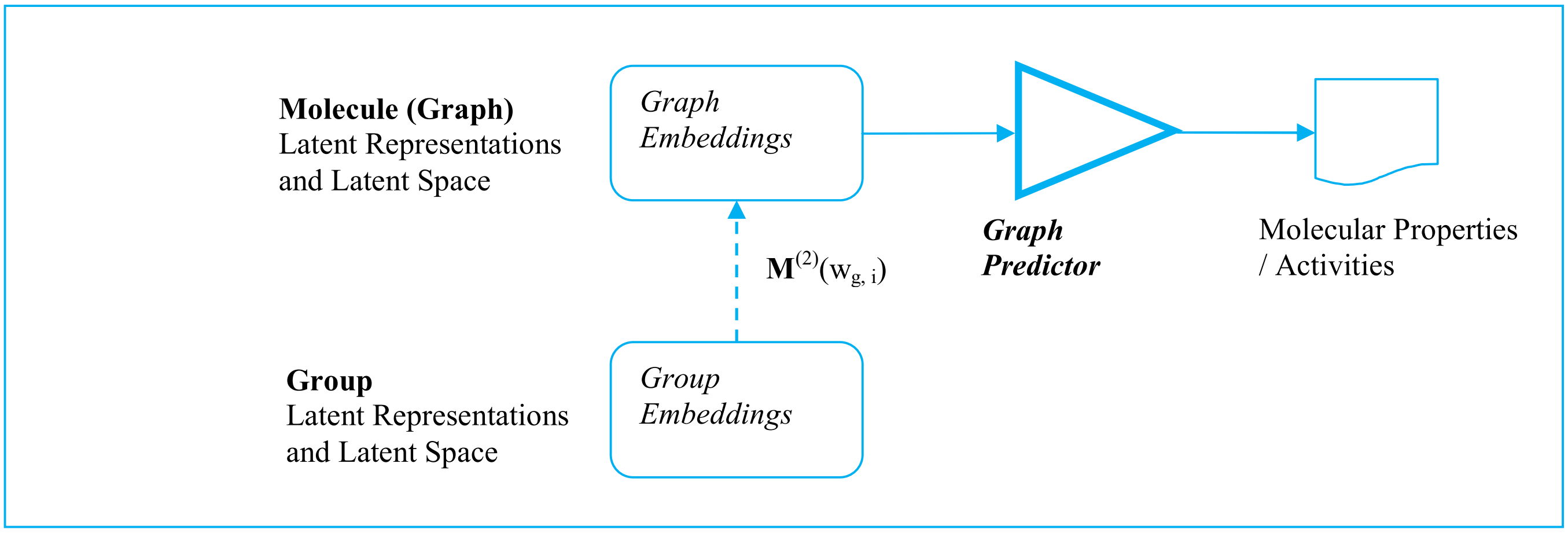

In the diagram, $\mathbf{M}^{(2)}$ is the *graph membership matrix*. Based on DiffGroupPool [1-2], we have

$$\mathbf{X}^{(3)} = (\mathbf{M}^{(2)})^{T}\mathbf{Z}^{(2)}$$

where $\mathbf{Z}^{(2)}$ is the group embeddings and $\mathbf{X}^{(3)}$ is the initial graph embeddings. $w_{g,\,i}$ is the *group weight* for group i in the graph membership matrix $\mathbf{M}^{(2)}$.

## 6.2 Weighted Group Pooling

To use tiered graph autoencoders with graph prediction we introduce the *group weight* $w_{g,\,i}$ [3] to represent and account for the different importance of group i for predicting different target molecular property / activity. Specifically we assign the group weight to a group i based on its type:

- $w_{g,\,i} = w_{FG}$ if i is a FG (functional group),
- $w_{g,\,i} = w_{RG}$ if i is a RG (ring group),
- $w_{g,\,i} = w_{CCG}$ if i is a CCG (connected-component group).

Using (variable) group weights $w_{g,\ i}$ in the graph membership matrix $\mathbf{M}^{(2)}$ amounts to *weighted group pooling* [3]. Weighted group pooling thus serves as the *link / coupling* between (unsupervised) tiered graph autoencoders and (supervised) graph prediction, enabling *task-independent node and group embeddings*, generated using tiered graph autoencoders, to be used to build *task-optimized graph embeddings*, generated using graph prediction for specific target molecular properties / activities.

## 6.3 Hyperparameter Optimization for Group Weights

To *optimize group weights* based on the results of graph prediction learning, we treat group weights as *hyperparameters* and use *hyperparameter optimization* [4-6] to accomplish it. Optimization of group weights is separate from learning of graph prediction, but depends on the latter's results.

In this approach [3], we use *graph embeddings*, generated using tiered graph autoencoders at Tier 3, as input for graph prediction (see 6.1 Tiered Graph Autoencoders with Graph Prediction). Note that graph embeddings depend on $\mathbf{M}^{(2)}$ and therefore on *group weights*. They need to be regenerated whenever a new configuration of group weights is selected in the hyperparameter optimization process. *Group embeddings (and node embedding)* are used in the generation of graph embeddings, but *need to be generated only once for a dataset*, independent of graph prediction and hyperparameter optimization, greatly reducing their computational cost and execution time.

For group weights optimization, we need to (1) select a *strategy* (e.g., Bayesian Optimization) for group weights (hyperparameters) search, (2) based on the strategy, design a *set, or sets, of group weights* to use, (3) generate *graph embeddings* for the designed set(s) of group weights, (4) perform a *target-specific graph prediction learning* for each generated graph embedding, (5) determine *the optimal set of group weights* that produce the best graph prediction results for the target, and (6) repeat from step 2, if the strategy calls for, until a *termination condition* is reached.

## 6.4 Group Weights Optimization Using Ax, BoTorch and GPyTorch

Based on earlier discussions on hyperparameter optimization, we select Bayesian optimization with Gaussian processes for group weights (hyperparameters) optimization. In particular, we select Ax, BoTorch and GPyTorch as the unified framework to use for Bayesian hyperparameter optimization.

Ax provides the *optimize*() function to construct and run a full *OptimizationLoop*, and to return the *best hyperparameter configuration*. We call the optimize() function with the following input parameters:

- hyperparameters: $w_{FG}$, $w_{RG}$, $w_{CCG}$
- evaluation function: train_evaluate
- minimize: True
- total trials (default: 20)

The evaluation function evaluates the *objective function* f(**x**) given a hyperparameter configuration **x**.

The evaluation function, *train_evaluate(parametrization)*, that we provide has three components:

1. tgae(parametrization)
2. train(training dataset)
3. evaluate(validation dataset)

The *hyperparameter configuration* (parametrization) is automatically generated by Ax for each Trial during a full run of the OptimizationLoop.

Given a hyperparameter configuration, the *tgae(parametrization)* function generates *graph embeddings* using tiered graph autoencoders and pre-generated / saved group embeddings, as discussed previously. The *train()* function trains the *graph prediction MLP*, given the training dataset that contains graph embeddings. The *evaluate()* function evaluates the validation dataset given the trained graph prediction MLP, and returns the *validation error* which serves as the *objective score*.

For the feasibility study we use the QM9 dataset [3] which has 133K drug-like organic molecules and 12 target molecular properties. Initially, we used *1000 samples* (80% training, 10% validation, 10% test) and *20 total trials* (default). For each target molecular property, we did two runs and the best hyperparameter configurations obtained are shown as the *first two results* in the table that follows. We then increased the sample size to *10000*. The best hyperparameter configuration obtained is shown as the *third result*. We finally also increased the number of *total trials to 40*. The best hyperparameter configuration obtained is shown as the *fourth result*. For each target molecular property, it corresponds to the *best hyperparameter configuration (in bold)* obtained by the feasibility study.

| Target Molecular Property | $w_{FG}$ | $w_{RG}$ | $w_{CCG}$ |
|---|---|---|---|
| mu (dipole moment) | 0.85<br>0.01<br>0.76<br>**0.86** | 0.26<br>0.70<br>0.69<br>**0.89** | 0.70<br>0.49<br>0.27<br>**0.31** |
| alpha (isotropic polarizability) | 0.79<br>0.87<br>0.97<br>**0.93** | 0.11<br>0.34<br>0.95<br>**0.11** | 0.37<br>0.10<br>0.90<br>**0.12** |
| HOMO (highest occupied molecular orbital energy) | 0.37<br>0.90<br>1.00<br>**0.07** | 0.03<br>0.57<br>0.32<br>**0.72** | 0.07<br>0.84<br>0.03<br>**0.80** |
| LUMO (lowest unoccupied molecular orbital energy) | 0.63<br>0.10<br>0.00<br>**0.00** | 0.89<br>0.76<br>0.67<br>**0.85** | 0.15<br>0.61<br>0.52<br>**0.21** |
| gap (gap between HOMO and LUMO) | 0.54<br>0.70<br>0.37<br>**0.29** | 0.73<br>1.00<br>1.00<br>**0.21** | 0.39<br>0.75<br>0.59<br>**0.34** |
| $R^2$ (electronic spatial extent) | 0.73<br>0.74<br>0.48<br>**0.31** | 0.65<br>0.72<br>1.00<br>**0.46** | 0.60<br>0.69<br>0.33<br>**0.21** |
| ZPVE (zero point vibrational energy) | 0.84<br>0.94<br>0.23<br>**0.93** | 0.82<br>0.76<br>0.00<br>**0.33** | 0.69<br>0.32<br>0.00<br>**0.06** |
| $U_0$ (internal energy at 0K) | 0.40<br>0.12<br>0.81<br>**0.19** | 0.67<br>0.35<br>0.76<br>**0.40** | 0.94<br>0.25<br>0.70<br>**0.64** |
| U (internal energy at 298.15K) | 0.10<br>0.80<br>0.65<br>**0.05** | 0.33<br>0.46<br>0.82<br>**0.97** | 0.90<br>0.07<br>0.58<br>**0.18** |
| H (enthalpy at 298.15K) | 0.75<br>0.84<br>0.44<br>**0.31** | 0.76<br>0.47<br>0.53<br>**0.64** | 0.41<br>0.77<br>0.42<br>**0.07** |
| G (free energy at 298.15K) | 0.42<br>0.43<br>0.81<br>**0.60** | 0.70<br>0.45<br>0.95<br>**0.34** | 0.10<br>0.74<br>0.09<br>**0.79** |
| Cv (heat capacity at 298.15K) | 0.98<br>0.64<br>0.66<br>**0.67** | 0.53<br>0.86<br>0.00<br>**0.54** | 0.03<br>0.97<br>0.80<br>**0.66** |

As an aside, we observe that different types of groups (*FG, RG, and CCG*) have different significance in predicting different molecular properties. Therefore, it is important to explicitly represent and utilize *groups and group weights* in deep learning for molecular graphs [1-3].

## 7 Summary and Conclusion

In this paper, we discussed Bayesian hyperparameter optimization, including hyperparameter optimization, Bayesian optimization, and Gaussian processes. We also reviewed BoTorch, GPyTorch and Ax, the new open-source frameworks that we used for Bayesian optimization, Gaussian process inference and adaptive experimentation, respectively.

For experimentation, we applied Bayesian hyperparameter optimization, for optimizing group weights, to weighted group pooling, which couples unsupervised tiered graph autoencoders learning and supervised graph prediction learning for molecular graphs.

We find that Ax, BoTorch and GPyTorch together provide a simple-to-use but powerful framework for Bayesian hyperparameter optimization, using Ax's high-level API that constructs and runs a full optimization loop and returns the best hyperparameter configuration.

As an aside, we find that different types of groups have different significance in predicting different molecular properties. Therefore, it is important to explicitly represent and utilize groups and group weights in deep learning for molecular graphs.

## References

[1] Daniel T. Chang, "Tiered Graph Autoencoders with PyTorch Geometric for Molecular Graphs," arXiv preprint arXiv:1908.08612 (2019).
[2] Daniel T. Chang, "Tiered Latent Representations and Latent Spaces for Molecular Graphs," arXiv preprint arXiv:1904.02653 (2019).
[3] Daniel T. Chang, "Deep Learning for Molecular Graphs with Tiered Graph Autoencoders and Graph Prediction," arXiv preprint arXiv:1910.11390 (2019).
[4] T. Hinz, N. Navarro-Guerrero, S. Magg, and S. Wermter, "Speeding Up the Hyperparameter Optimization of Deep Convolutional Neural Networks," in International Journal of Computational Intelligence and Applications Vol. 17, No. 2 (2018).
[5] Charlie Harrington, "Practical Guide to Hyperparameters Optimization for Deep Learning Models," 2018 (https://blog.floydhub.com/guide-to-hyperparameters-search-for-deep-learning-models/).
[6] Will Koehrsen, "A Conceptual Explanation of Bayesian Hyperparameter Optimization for Machine Learning," 2018 (https://towardsdatascience.com/a-conceptual-explanation-of-bayesian-model-based-hyperparameter-optimization-for-machine-learning-b8172278050f).
[7] Jasper Snoek, Hugo Larochelle, and Ryan P. Adams, "Practical Bayesian Optimization of Machine Learning Algorithms," in Advances in Neural Information Processing Systems, pp. 2951–2959, 2012.
[8] Bobak Shahriari, Kevin Swersky, Ziyu Wang, Ryan P Adams, and Nando de Freitas. "Taking the Human Out of the Loop: A Review of Bayesian Optimization," in Proceedings of the IEEE, 104(1): 148–175, 2016.
[9] Peter I. Frazier, "A Tutorial on Bayesian Optimization," arXiv preprint arXiv:1807.02811 (2018).
[10] Ian Dewancker, Michael McCourt, and Scott Clark, "Bayesian Optimization Primer," 2015 (https://sigopt.com/static/pdf/SigOpt_Bayesian_Optimization_Primer.pdf).
[11] I. Dewancker, M. McCourt, and S. Clark, "Bayesian Optimization for Machine Learning: A Practical Guidebook," arXiv preprint arXiv:1612.04858 (2016).
[12] C. E. Rasmussen and C. K. Williams, *Gaussian Processes for Machine Learning*, MIT Press, 2006.

[13] Haitao Liu, Yew-Soon Ong, Xiaobo Shen, and Jianfei Cai, "When Gaussian Process Meets Big Data: A Review of Scalable GPs," arXiv preprint arXiv:1807.01065 (2018).
[14] Maximilian Balandat, Brian Karrer, Daniel R. Jiang, Samuel Daulton, Benjamin Letham, Andrew Gordon Wilson, and Eytan Bakshy. "BoTorch: Programmable Bayesian Optimization in PyTorch," arXiv preprint arXiv: 1910.06403 (2019).
[15] BoTorch. Bayesian Optimization in PyTorch (https://botorch.org/).
[16] J. R. Gardner, G. Pleiss, D. Bindel, K. Q. Weinberger, and A. G. Wilson, " GPytorch: Blackbox Matrix-matrix Gaussian Process Inference with GPU Acceleration," in 32nd Conference on Neural Information Processing Systems (NIPS 2018).
[17] GPyTorch. Gaussian Processes for Modern Machine Learning Systems (https://gpytorch.ai/).
[18] Eytan Bakshy, Lili Dworkin, Brian Karrer, Konstantin Kashin, Benjamin Letham, Ashwin Murthy, and Shaun Singh, "AE: A Domain-agnostic Platform for Adaptive Experimentation," in SemanticScholar (2018).
[19] Eytan Bakshy, Max Balandat, and Kostya Kashin, "Open-sourcing Ax and BoTorch: New AI tools for Adaptive Experimentation" (https://ai.facebook.com/blog/open-sourcing-ax-and-botorch-new-ai-tools-for-adaptive-experimentation/).
[20] Ax. Adaptive Experimentation Platform (https://ax.dev/).